\documentclass[sigconf]{acmart}

\pdfoutput=1

\usepackage{booktabs} 

\usepackage{wrapfig}
\usepackage{ifthen}
\usepackage{makecell}

\setcopyright{rightsretained}

\acmConference[DEBS'18]{12th ACM International Conference on Distributed and Event-Based Systems}{June 2018}{Hamilton, New Zealand}
\acmYear{2018}
\copyrightyear{2018}

\newboolean{showcomments}
\setboolean{showcomments}{true}
\ifthenelse{\boolean{showcomments}}
{ \newcommand{\mynote}[3]{
    \fbox{\bfseries\sffamily\scriptsize#1}
    {\small$\blacktriangleright$\textsf{\emph{\color{#3}{#2}}}$\blacktriangleleft$}}}
{ \newcommand{\mynote}[3]{}}

\begin{document}
\title{Grand Challenge: Cell Grid Architecture for Maritime Route Prediction on AIS Data Streams}
\renewcommand{\shorttitle}{Cell Grid Architecture for Maritime Route Prediction on AIS Data Streams}

\author{Ciprian Amariei}
\affiliation{%
  \institution{Alexandru Ioan Cuza University of Ia\c{s}i, Romania}
}
\email{ciprian.amariei@gmail.com}

\author{Paul Diac}
\affiliation{%
  \institution{Alexandru Ioan Cuza University of Ia\c{s}i, Romania}
}
\email{paul.diac@info.uaic.ro}

\author{Emanuel Onica}
\affiliation{%
  \institution{Alexandru Ioan Cuza University of Ia\c{s}i, Romania}
}
\email{eonica@info.uaic.ro}

\author{Valentin Ro\c{s}ca}
\affiliation{%
  \institution{Alexandru Ioan Cuza University of Ia\c{s}i, Romania}
}
\email{valentin.rosca@info.uaic.ro}

\begin{abstract}
The 2018 Grand Challenge targets the problem of accurate predictions on data streams produced by automatic identification system (AIS) equipment, describing naval traffic. 
This paper reports the technical details of a custom solution, which exposes multiple tuning parameters, making its configurability one of the main strengths. 
Our solution employs a cell grid architecture essentially based on a sequence of hash tables, specifically built for the targeted use case.
This makes it particularly effective in prediction on AIS data, obtaining a high accuracy and scalable performance results.
Moreover, the architecture proposed accommodates also an optionally semi-supervised learning process besides the basic supervised mode.
\end{abstract}

\ccsdesc[500]{Computing methodologies~Machine learning algorithms}
\ccsdesc[300]{Information systems~Data structures}
\ccsdesc[300]{Applied computing~Transportation}

\keywords{data stream processing, machine learning, event based systems}

\copyrightyear{2018} 
\acmYear{2018} 
\setcopyright{acmcopyright}
\acmConference[DEBS '18]{The 12th ACM International Conference on Distributed and Event-based Systems}{June 25--29, 2018}{Hamilton, New Zealand}
\acmBooktitle{DEBS '18: The 12th ACM International Conference on Distributed and Event-based Systems, June 25--29, 2018, Hamilton, New Zealand}
\acmPrice{15.00}
\acmDOI{10.1145/3210284.3220503}
\acmISBN{978-1-4503-5782-1/18/06}

\maketitle

\section{Introduction}
\label{sec:intro}

Accurate predictions on automatic identification system (AIS) data streams produced by vessels can be an important factor in optimizing supply chain management in the maritime transportation.
The 2018 DEBS Grand Challenge~\cite{GC2018} focuses on predicting destination of ships and arrival times.
The context given is of routes between a set of ports in the Mediterranean Sea.
The training data sets contain AIS tuples emitted by ships, labeled with their destination and arrival times, while the evaluation data is not labeled.
We will refer throughout this paper at the various information included in an AIS tuple (course, speed, etc.) as the AIS tuple~\emph{dimensions}.

The central idea of our solution is to split the data according to a cell grid corresponding to the monitored geographical space.
The training and prediction is then executed on a \emph{per cell} basis, where the cell dimension is a configurable property of the cell grid.

Our solution is composed of the following main execution stages:

\begin{itemize}
\item The initial supervised learning: training data generated in each cell is aggregated and stored in a specific data structure, essentially a sequence of hash tables maintained per cell.
\item The prediction: each new unlabeled AIS tuple is matched over the structures in the cell where the tuple originated from, finally obtaining the best candidate.
\item The optional robustness test: an evaluation is done to determine if the best candidate prediction does not affect the robustness of the prediction stream.
\end{itemize}

The destinations prediction follows all the three stages above, while predicting the arrival times just the first two.
The solution also permits executing a semi-supervised learning during the prediction stages.
We describe in Section~\ref{sec:solution} the details of the three stages and the architecture of the used data structures.
Finally, we discuss some obtained results and performance consideration in Section~\ref{sec:conclusion}.

\section{Solution Architecture}
\label{sec:solution}

We describe in the following the stages executed by our solution for each of the two types of predictions. 

\subsection{Predicting Destinations}

\begin{figure}
\includegraphics[width=0.40\textwidth]{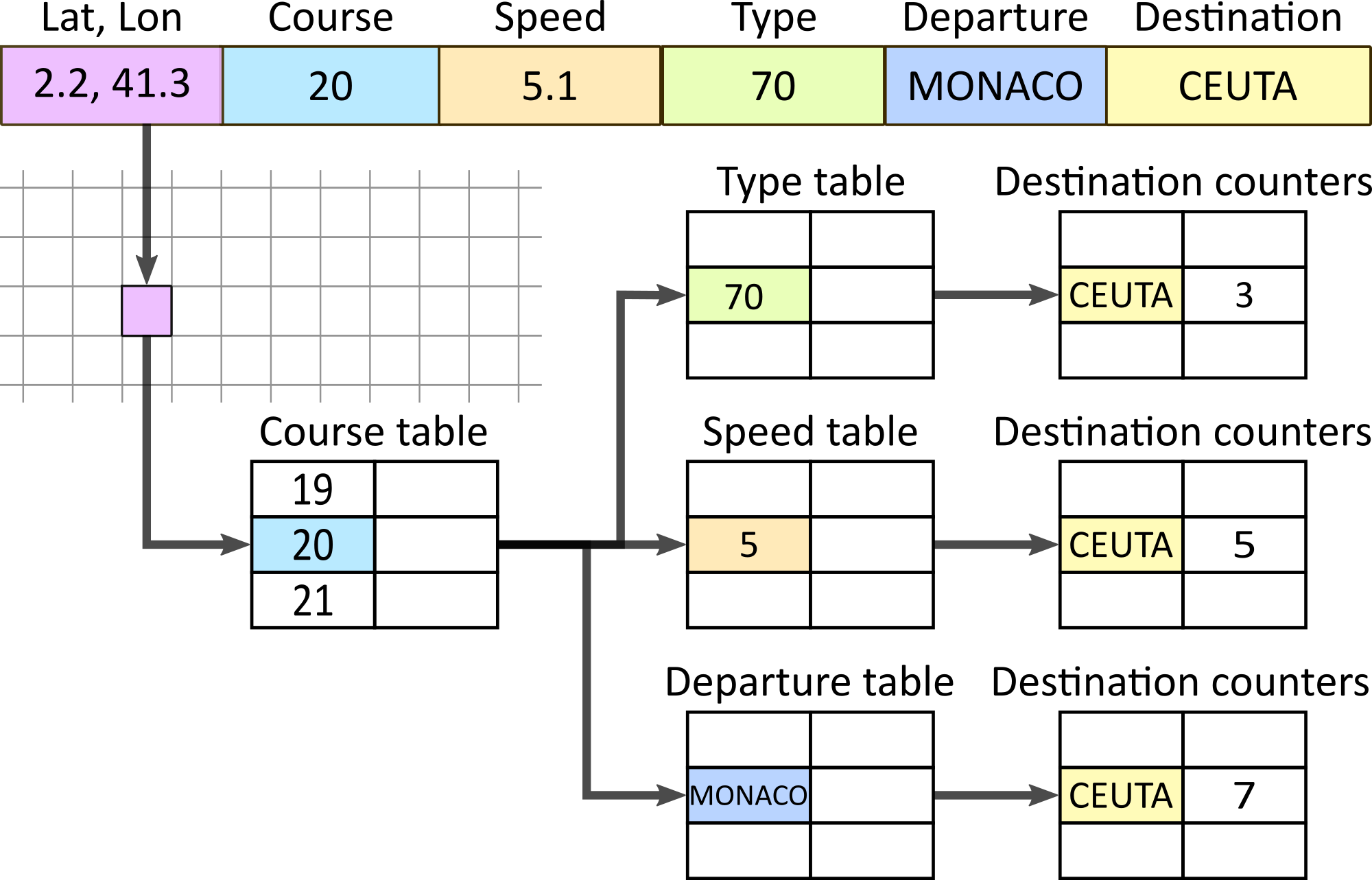}
\centering
\caption{Data structures used for destination prediction.}
\label{fig:archq1}
\vspace{-15pt}
\end{figure}

\emph{The initial supervised learning stage:}
We present an example showing the learning and data structures maintained per cell for destination prediction in Figure~\ref{fig:archq1}.
The first step is to compute the corresponding cell of the training AIS tuple based on geographical coordinates.
Afterwards, AIS tuple data is indexed in a sequence of hash tables maintained for the cell - we call this \emph{the training of the cell}.

The key in the first hash table is given by the course dimension of the AIS tuple.
For each populated course key, the stored value consists in three more separate hash tables keyed on ship type, speed\footnote{The speed values, being in a continuous domain, are discretized using a configurable granularity, which results in speed intervals.} and departure port.
Finally, the values stored in each of the three maps, are represented by a set of destination ports counters.
Each learned AIS tuple will increment three counters of its destination port as presented in the Figure~\ref{fig:archq1} example (in this case the CEUTA counters will increase).

\begin{figure}[h]
\includegraphics[width=0.40\textwidth]{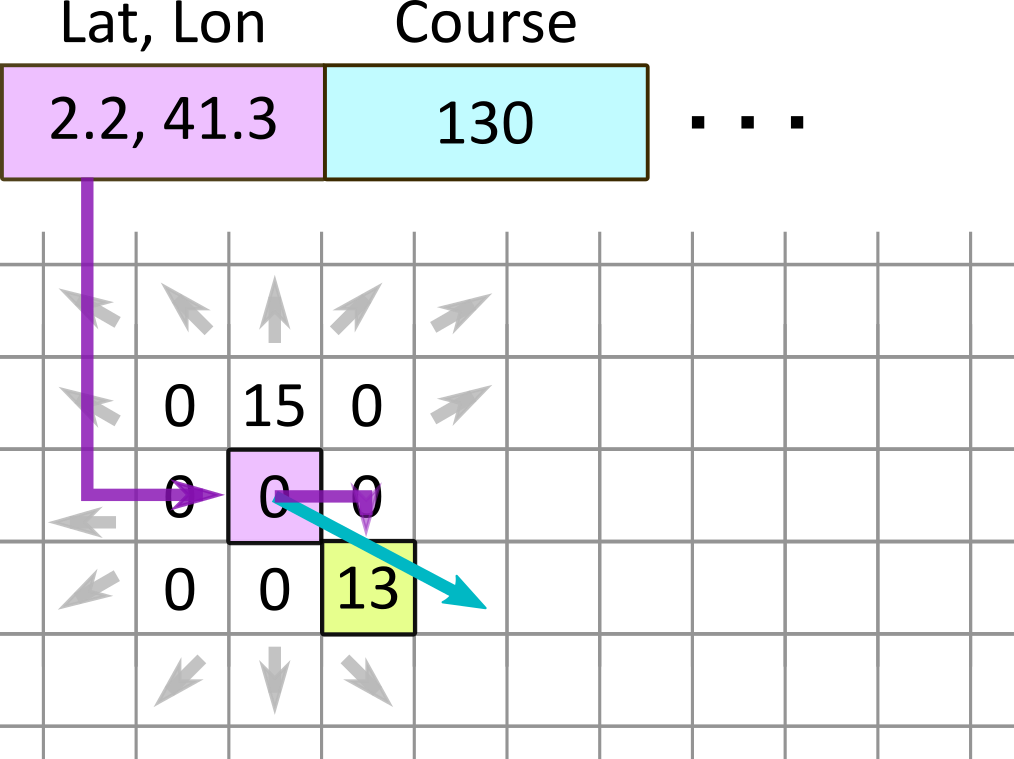}
\centering
\caption{Finding the closest trained cell.}
\label{fig:closest}
\end{figure}

\smallskip
\noindent
\emph{The prediction stage:}
The received unlabeled AIS tuples are matched over the hash table sequence in the cell where they were generated.
If the cell was not trained, then an efficient algorithm searches for the closest best fit trained cell.
This is done by gradually searching the surrounding cells. 
The chosen cell is the one which had the highest number of AIS tuples learned and fits closest with the course of the evaluated AIS tuple.
We display in Figure~\ref{fig:closest} a relevant example of this situation (the numbers in the cells identify the frequency of trained tuples for the respective cells).
Because the course in the evaluated AIS tuple is 130 the cell in the south-east is preferred as most relevant.
If this would be untrained the north cell would be chosen.

For matching, first, we compute the course approximation to select the closest course keys.
The matching continues for each of the selected course keys, over the three corresponding separate hash tables. 
For each of these dimensions that matches the evaluated AIS tuple, we select the corresponding destination counters sets as \emph{candidates}.

The reason of prioritizing course is because we observed that this is one of the dimensions that correlates better with destinations.
For determining this, we have used the training set both for training and evaluation in some preliminary tests.
Essentially, by simply selecting a prediction only using a single dimension counter we have obtained the results displayed in Table~\ref{tab:errorrates}.
We have decided to use the course as a first level of matching, also because we can establish a course tolerance for selecting multiple candidate matches (an alternative option is using the departure port).
Although it seems that the heading is on par with the course in terms of error rate, some of the training tuples were missing the heading information. 
Therefore, we preferred not to rely on this dimension.

\begin{table}[h]
\begin{center}
 \begin{tabular}{||c c||}
 \hline
 \thead{\textbf{Dimension}} & \thead{\textbf{Error rate percentage}}\\ 
 \hline\hline
 Type & 41\% \\
 \hline
 Speed & 42.5\% \\
 \hline
 Departure & 30.5\% \\
 \hline
 Draught & 45\% \\
 \hline
 Course & 38\% \\
 \hline
 Heading & 38\% \\
 \hline
\end{tabular}
\vspace{5pt}
\caption{Error rate in observing correlations with destination ports}
 \label{tab:errorrates}
\end{center}
\end{table}

\vspace{-15pt}


Following, we perform a sequence of aggregation operations over the selected destination counters candidate sets to determine the best prediction.
The dominant factor in choosing a best candidate is the counters sum per specific candidate destination, but other criteria can also be applied.
These include geographical criteria (e.g., closest port by course or distance to the evaluated tuple), the destination port having the most arrivals from the departure port in the evaluated AIS tuple, or the destination port having the most arrived ships of a similar type with the one in the evaluated AIS tuple.


\smallskip
\noindent
\emph{The optional robustness test:}
If the chosen best candidate is different from the previous destination prediction for the same ship, this means that the new prediction is uncertain.
Therefore, in a final step, we evaluate if the best destination prediction candidate is reliable enough to be actually reported as result.
The main idea is to observe if the predicted destination is part of the longest contiguous row of predictions of the same port. 
In this case, the current detected prediction is also the prediction reported.
Otherwise, it is preferred to report as prediction the port corresponding to the dominant longest contiguous row.
The algorithm uses several configurable prediction frequency statistics for fine-tuning it in different flavors (e.g., considering multiple k-longest consistent prediction rows).
This significantly increases the robustness of the predictions stream. 

\subsection{Predicting Arrival Time}

The prediction of arrival time requires the result obtained as destination port prediction.
Another cell grid is used, with finer cell granularity.  
This is needed to obtain accurate results.

\smallskip
\noindent
\emph{The initial supervised learning stage:}
The arrival times information in the labeled tuples is stored using a slightly different sequence of hash tables maintained per trained cell.
The first key to index the information is the destination port.
The value in the destination table is represented as before by a set of multiple separate hash tables indexed on the various dimensions in the AIS tuple\footnote{One main difference to the structure used in the destination prediction is that any of the dimension hash tables is optional}. 
For each dimension key in these second-level tables, the final values are a set of arrival time statistics, including the average time taken from the cell to the destination port. 
A similar set of global statistics is kept 
for all ships reaching a key destination port.
For each time statistics kept, also the corresponding \emph{reference AIS tuples} closest to the average time are preserved.

\smallskip
\noindent
\emph{The prediction stage:}
We perform the matching for the evaluated AIS tuple similarly to the destination prediction stage.
The architecture configuration permits selecting a dimension criteria in the second-level maps on which to give the average time prediction.
We observed that it is sufficient to use the statistics gathered for a single dimension to obtain good predictions.
It seems the best dimensions are course, departure and speed providing relatively close results.

\begin{figure}[h]
\includegraphics[width=0.40\textwidth]{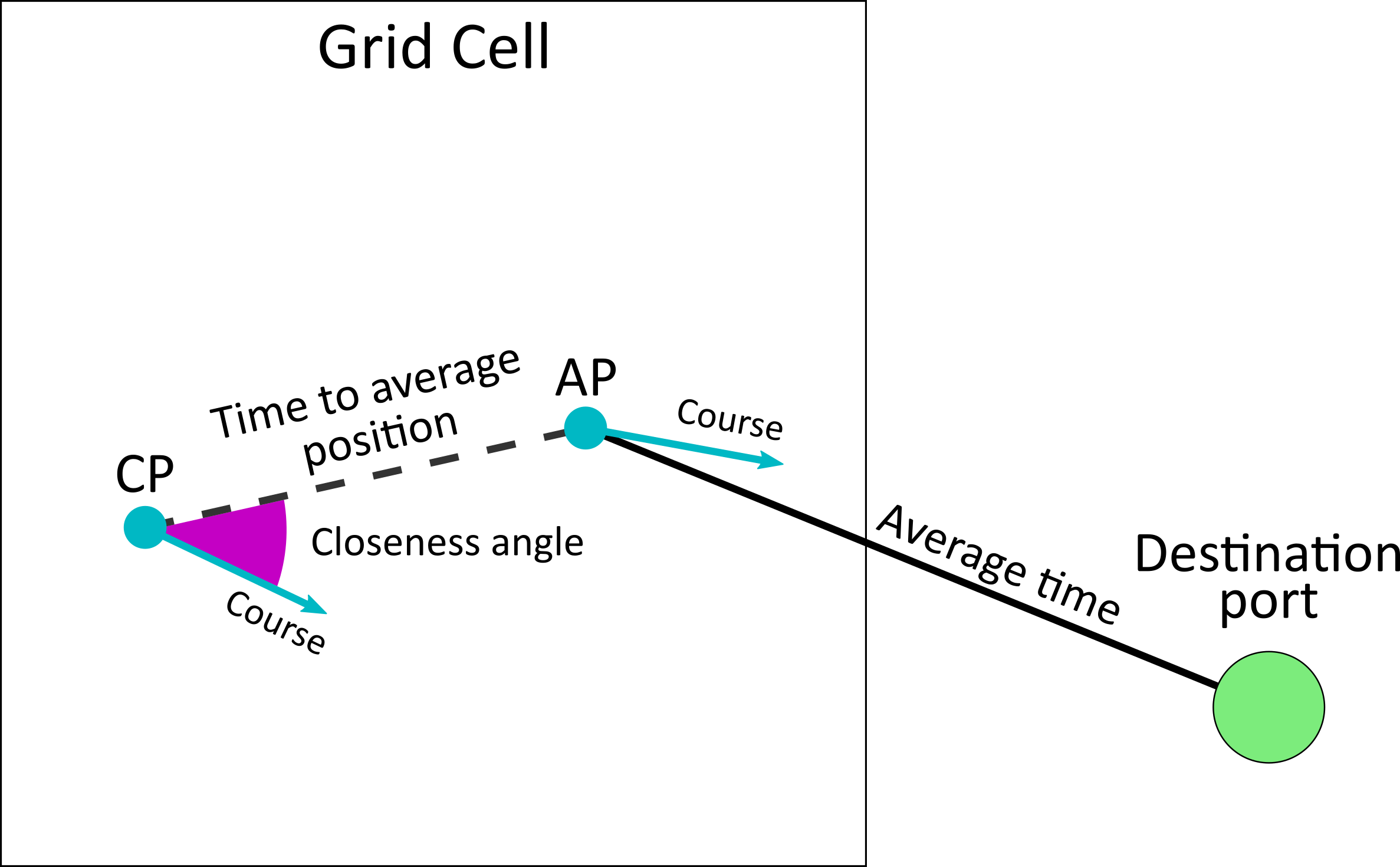}
\centering
\caption{Improving the time estimation error.}
\label{fig:timeestimation}
\end{figure}

An interesting improvement in the time estimation error can be obtained at the cell level using an approximation of the time that would be needed to sail the distance from the current evaluated AIS tuple to the reference AIS tuple recorded as closest to average time.
An example is depicted in Figure~\ref{fig:timeestimation}.

We compute if the ship that emitted the currently evaluated tuple (CP) is either before or after the reference AIS tuple (AP) in the cell, on its way towards destination.
In the given example it is found that the ship that emitted CP most probably trails behind the reference AIS tuple.
In this case, we compute the time it would take the ship to reach the position of the reference AIS tuple.
We add this time to the average time of getting from the cell to the destination port, obtaining the arrival time prediction.
We could proceed in a similar manner by subtracting the time between CP and AP out of the average time, if CP would be in front of the reference AP.
This time adjustment operation has also a degree of uncertainty and in some cases can increase the error.
We have found, however, for the given training sets, that in general a more accurate value can be obtained.


%
%

\subsection{Optional semi-supervised learning}

Our solution can execute also a semi-supervised learning during the prediction stage.
For this, the received unlabeled AIS tuples emitted by each ship are temporarily retained until the ship ends its current travel.
At the respective moment, the tuples emitted by a ship are labeled with the last predicted value for the destination.  
The arrival time is labeled with the time reported by the first AIS tuple that entered the radius of the destination port.
The difficulty lies in detecting correctly when a ship ends its trip. 
Currently, the solution we have implemented starts a short-lived thread tasked with detecting if a ship ends its trip whenever a ship starts reporting AIS tuples within a port radius.
If, after a while, the ship stops emitting tuples while still in the port radius, we conclude that the trip ended there, and the thread labels the trip tuples and executes the learning over them.
Otherwise, if the ship continues to emit tuples outside the port radius, it means that it is probably continuing its trip and the thread ends.


\section{Discussion and conclusion}
\label{sec:conclusion}

With configurations executed by the time of writing this paper, we predicted correctly the destination for 86 percent of the routes length with a mean error of 292 minutes on the arrival time.

We used a cell granularity of 1 degree for predicting destinations and 0.005 degrees for predicting times.
The number of cells might seem very high in the second case with respect to memory consumption: over 25 million with the grid dimensions adapted to the Mediterranean Sea use case.
However, we have observed that on average for all training sets just around 1 percent of these were trained storing data and actually used in the evaluation (in the range of 250000), which we consider an acceptable amount. 

In respect to the semi-supervised learning, the extra short-lived threads and periodical necessary synchronization, do not seem to affect significantly the running time of the solution.
However, we also did not observe any significant increase in accuracy, which we believe to be due to the relatively low number of AIS tuples evaluated compared to the size of the initial training sets.

We believe that one of the main strengths of our architecture is its extensive configurability.
We have obtained good results with the settings we have tried, but at the moment of writing of this paper, we believe that exploring the complete potential of our solution is still far.
Trying to adjust the variety of the exposed parameters by following more accurate preliminary data analysis or employing some automated solution based on genetic algorithms might provide even better results.

\section*{Acknowledgements}
\setlength{\columnsep}{9pt}
\setlength{\intextsep}{3pt}
\begin{wrapfigure}{r}{0.09\textwidth}
\includegraphics[width=\linewidth]{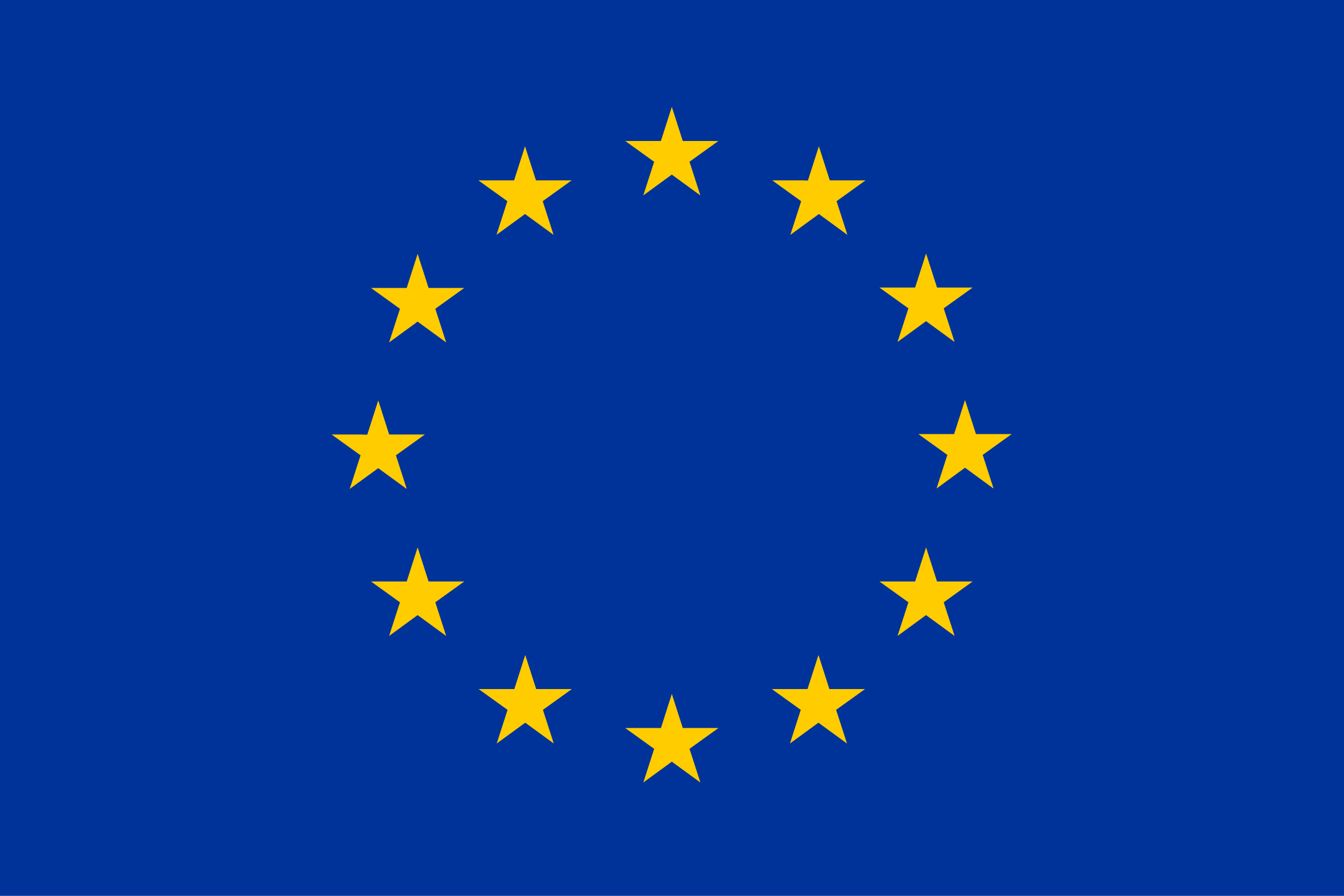}
\end{wrapfigure}
The dissemination of this work is partly funded by the \emph{European Union{\textquotesingle}s Horizon 2020 research and innovation programme} under grant agreement No 692178. This work was also partly supported by a grant of the Romanian National Authority for Scientific Research and Innovation, CNCS/CCCDI - UEFISCDI, project number 10/2016, within PNCDI III.

\bibliographystyle{ACM-Reference-Format}
\bibliography{references}

\end{document}